\documentclass[journal]{IEEEtran}
\usepackage{cite}

\usepackage{wrapfig} 
\usepackage{epsfig}
\usepackage{graphicx}

\usepackage{amsmath}
\usepackage{amssymb}
\usepackage{amsfonts}       
\usepackage{array}
\usepackage{url}

\usepackage[dvipsnames,svgnames,x11names]{xcolor}
\definecolor{citecolor}{RGB}{119,185,0} 
\usepackage[pagebackref=false,breaklinks=true,letterpaper=true,colorlinks,citecolor=citecolor,bookmarks=false]{hyperref}
\usepackage{soul}
\usepackage[utf8]{inputenc}
\usepackage{multirow}
\usepackage{comment}

\def\eg{\emph{e.g.}} 
\def\ie{\emph{i.e.}} 
\def\etal{\emph{et~al.}}

\newlength\savewidth\newcommand\shline{\noalign{\global\savewidth\arrayrulewidth
  \global\arrayrulewidth 1pt}\hline\noalign{\global\arrayrulewidth\savewidth}}

\usepackage{floatrow} 
\begin{document}

\title{Parameter-Efficient Person Re-identification \\ in the 3D Space}
\author{Zhedong Zheng, Nenggan Zheng, and Yi Yang,~\IEEEmembership{Senior~Member,~IEEE} \thanks{Zhedong Zheng and Yi Yang are with ReLER Lab, Australian Artificial Intelligence Institute, University of Technology Sydney, NSW 2007, Australia. E-mail: zdzheng12@gmail.com, yi.yang@uts.edu.au }
\thanks{
Nenggan Zheng is with the School of Computer Science, Zhejiang University, Hangzhou 310027, China. E-mail: zng@cs.zju.edu.cn }
}

\markboth{Journal of \LaTeX\ Class Files,~Vol.~14, No.~8, August~2015}%
{Shell \MakeLowercase{\textit{et al.}}: Bare Demo of IEEEtran.cls for IEEE Journals}

\maketitle

\begin{abstract}
 People live in a 3D world. However, existing works on person re-identification (re-id) mostly consider the semantic representation learning in a 2D space, intrinsically limiting the understanding of people. 
 In this work, we address this limitation by exploring the prior knowledge of the 3D body structure. 
 Specifically, we project 2D images to a 3D space and introduce a novel parameter-efficient Omni-scale Graph Network (OG-Net) to learn the pedestrian representation directly from 3D point clouds.
 OG-Net effectively exploits the local information provided by sparse 3D points and takes advantage of the structure and appearance information in a coherent manner.  With the help of 3D geometry information, we can learn a new type of deep re-id feature free from noisy variants, such as scale and viewpoint. To our knowledge, we are among the first attempts to conduct person re-identification in the 3D space.  
 We demonstrate through extensive experiments that the proposed method (1) eases the matching difficulty in the traditional 2D space, (2) exploits the complementary information of 2D appearance and 3D structure,  (3) achieves competitive results with limited parameters on four large-scale person re-id datasets, 
 and (4) has good scalability to unseen datasets. 
 Our code, models and generated 3D human data are publicly available at \url{https://github.com/layumi/person-reid-3d}.

\end{abstract}

\begin{IEEEkeywords}
Person re-identification, 3D human representation, Image retrieval, Point cloud,  Graph convolutional networks.
\end{IEEEkeywords}

\section{Introduction}\label{sec:introduction}
\IEEEPARstart{P}{erson} re-identification is usually regarded as an image retrieval problem of spotting the person in non-overlapping cameras \cite{gong2014re,zhang2016learning,zheng2016survey,zheng2018discriminatively,ye2020deep,wang2019beyond}. Due to the rising demand of public safety and the fast development of camera network, person re-id has received increasing interests. These studies aim to save the human resource and efficiently find the person of interest, \eg, lost child in the airport, from thousands of candidate images. 
In recent years, the advance of person re-id is mainly due to two factors: 1) the availability of large-scale datasets and 2) the deeply-learned person representation. On one hand, deeply-learned models are usually data-hungry. The large-scale datasets~\cite{zheng2015scalable,liu2016large,zheng2017unlabeled,wei2018person} facilitate the data-driven approaches. 
On the other hand, the development of Convolutional Neural Network (CNN) also provides the technical breakthrough of the pedestrian representation learning. Many efforts have been paid to improve the CNN-based model capability~\cite{zhou2019osnet,li2018harmonious,qian2017multi,qian2019leader}. Recently, some researchers and companies also claim that the model can surpass the human performance \cite{zhang2017alignedreid}.

However, one inherent problem still remains: does the model really understand the person? People live in a 3D world. In contrast, we notice that most prevailing person re-id methods ignore the prior knowledge that human is a 3D non-rigid object, and only focus on learning the representation in 2D space. 
Although some pioneering works~\cite{barbosa2018looking,sun2019dissecting} consider the 3D human structure, the pedestrian representation is still learned from the projected 2D images. For instance, one of the existing works, PersonX \cite{sun2019dissecting}, has applied the game engine to build 3D person models. However, representation learning is conducted in the 2D space by projecting the 3D model back to 2D images. This line of works is effective in data augmentation but might be sub-optimal in representation learning. It is because the 2D data space intrinsically limits the model to understand the 3D geometry information of the person.

\begin{figure}[t]
\begin{center}
     \includegraphics[width=1\linewidth]{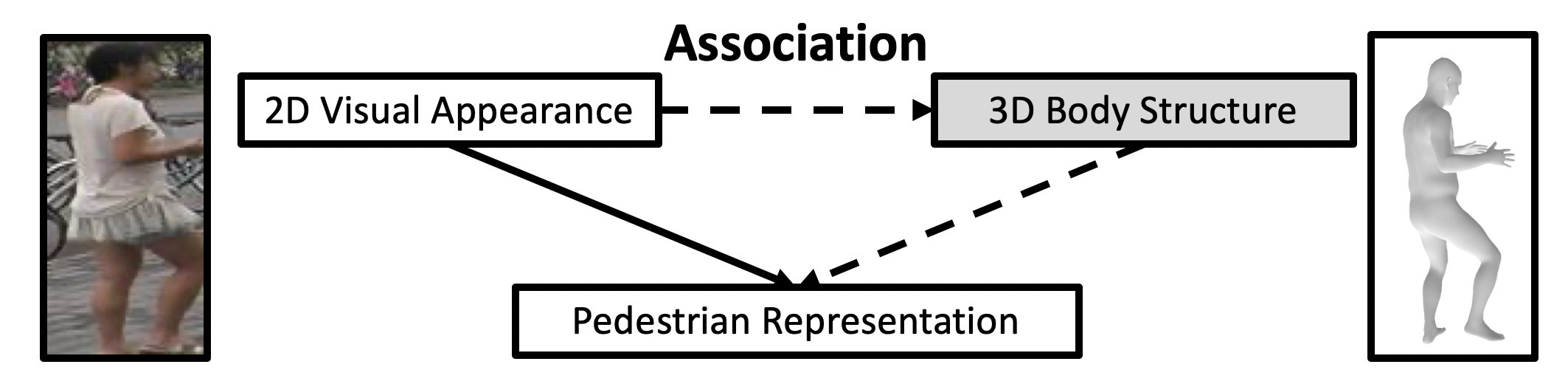}
\end{center} 
      \caption{ 
      Our brain generally associates the 2D appearance with prior knowledge of the 3D body shape. In this work, we intend to simulate this process and explore robust pedestrian representation with a lightweight model. (Dash arrows are missing in prevailing re-id methods.)
      }
      \label{fig:motivation}
\end{figure}

\begin{figure*}[t]
\begin{center}
     \includegraphics[width=1\linewidth]{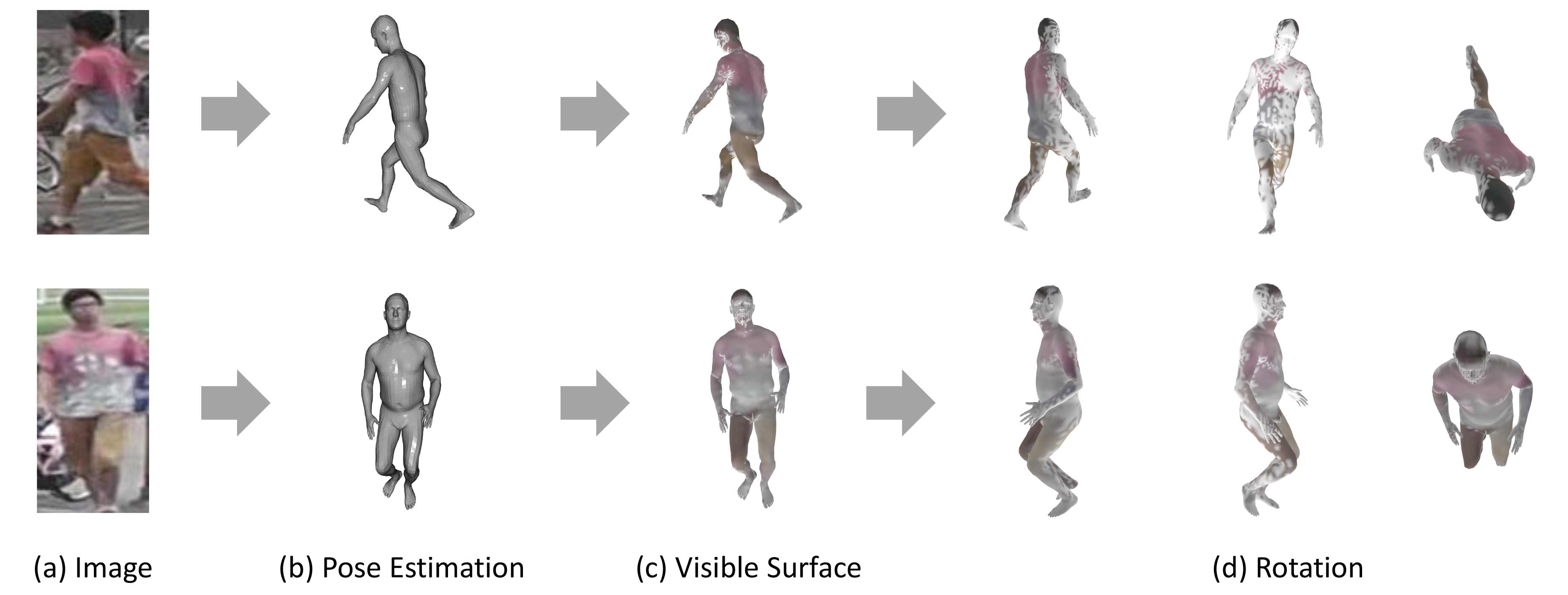}
\end{center} 
      \caption{ Person is a 3D non-rigid object. In this work, we conduct the person re-identification in the 3D space, and learn a new type of robust re-id feature. Given one 2D image \textbf{(a)}, we first \textbf{(b)} estimate the 3D pose via the off-the-shelf model \cite{kanazawaHMR18}, followed by \textbf{(c)} mapping the RGB color of visible surfaces to corresponding points. The invisible parts are made transparent for visualization. \textbf{(d) The appearance information is aligned with the human structure. We make the person free from the 2D space, and thus ease the matching difficulty.} }
      \label{fig:data}
\end{figure*}

Inspired by the human ability of associating the 2D appearance with the 3D geometry structure (see Figure~\ref{fig:motivation}), we argue that the key to learning an effective and scalable person representation is to consider the complementary information of 2D human appearance and 3D geometry structure. With the prior knowledge of 3D human geometry information, we could learn a depth-aware model, thus making the representation robust to real-world scenarios. As shown in Figure~\ref{fig:data}, we map the visible surface to the human mesh, and make the person free from the 2D space. 
The intuition is that after mapping to the 3D space, the appearance information is correlated/aligned with the human structure.
Without the need to worry about the part matching from two different viewpoints, the 3D data structure eases the matching difficulty in nature. The model could concentrate on learning the identity-related features, and dealing with the other intra-class variants, such as illumination conditions. 

To fully take advantage of the 3D structure and 2D appearance, we propose a novel Omni-scale Graph Network for person re-id in the 3D space, called OG-Net. OG-Net is a parameter-efficient model based on graph neural network (GNN) to communicate between the discrete cloud points of arbitrary locations. Given the 3D point cloud and the corresponding color information, OG-Net predicts the person identity and outputs the robust human representation for subsequent matching. 
Following the spirit of the conventional convolutional neural network (CNN), we utilize 3D points to build the location topology, and deploy the corresponding RGB color to extract appearance information.  In particular, we propose Omni-scale module to aggregate the feature from multiple 3D receptive fields, which leverages multi-scale information in 3D data. 
Even though the basic OG-Net only consists of four Omni-scale modules, it has achieved competitive performance on four person re-id datasets. 

\textbf{Contribution.} Our contributions are as follows. 
(1) We study person re-identification in the 3D space - a realistic scenario which could better reflect the nature of the 3D non-rigid human. To our knowledge, this work is among the early attempts to address this problem. (2) We propose a novel Omni-scale Graph Network to learn the feature from both human appearance and 3D geometry structure in a coherent manner. OG-Net leverages discrete 3D points to capture the multi-scale identity information. (3) Extensive experiments on four person re-id benchmarks show the proposed method could achieve competitive performance with limited parameters. 
A more realistic transfer learning setting is also studied in this paper. We observe that OG-Net has good scalability to the unseen person re-id dataset. 

\section{Related work}
\subsection{Semantic Space for Person Re-id} 
Recent years, convolutional neural network (CNN) models have been explored to map the pedestrian inputs, \eg, images, into one shared semantic space, where the data of the same identity is close and the data of different identities is apart from each other~\cite{yang2017enhancing,zhang2016learning}. Different optimization objectives have been studied. For instance, the contrastive loss is widely-used to discriminate different identities \cite{yi2014deep,zheng2018discriminatively,lin2018unsupervised}, while the identification loss deploys the identity classification as the pretext task~\cite{zheng2016survey,zhong2018camera,yu2017devil}. To simultaneously minimize the intra-class difference and maximize the inter-class gap, the triplet loss with different hard sampling strategies are also widely-studied \cite{hermans2017defense,ristani2018features,yang2018person}. Xiao \etal~\cite{xiao2017joint} propose the online instance matching loss to view the unlabeled data as negative samples, while Zheng \etal~\cite{zheng2017unlabeled} design one label smooth loss to take advantage of synthetic data. Besides, several works~\cite{lin2019improving,wang2018transferable,wang2017attribute} utilize person attributes, \eg, gender, to help the model learning intermediate features. This line of works is orthogonal to our work - any semantic spaces or optimization objectives can be used in our work and better ones can benefit our approach. In this work, we do not intend to pursue the best semantic space, but focus on verifying the effectiveness of the 3D space and the proposed OG-Net. We, therefore, deploy the basic identification loss for a fair comparison.

\subsection{Part Matching for Person Re-id}
To obtain the discriminative pedestrian representation, one line of research works resorts to mining local patterns, such as bodies, legs and arms, on 2D image inputs. The part matching is usually conducted on two different levels, \ie, the pixel level~\cite{su2017pose,zhang2019densely,zheng2019pose} and the feature level~\cite{sun2017beyond,wang2018learning,wu2020deep,hou2020iaunet}. 
The pixel-level part matching directly transforms the input image to one unified form. For instance, Su \etal~\cite{su2017pose} and Zheng \etal~\cite{zheng2019pose} deploy the off-the-shelf pose estimator~\cite{wei2016convolutional} to predict the human key points, followed by cropping and resizing body parts for representation learning. Similarly, Zhang \etal~\cite{zhang2019densely} utilize the semantic segmentation predictor to crop and align body parts densely. Instead of cropping body parts, Saquib \etal~\cite{saquib2018pose} concatenate the rgb input with key point heatmap as input, and let model to learn the part attention by itself.
In contrast, another line of works align the parts coarsely on the feature level, given that pedestrians usually stand in the image and are horizontally aligned in nature. Based on this assumption, Sun \etal~\cite{sun2017beyond,sun2019learning} propose to split feature 
maps horizontally and learn the part feature in a relatively large receptive field. Taking one more step, MGN~\cite{wang2018learning} explores more partition strategies and fuses different loss functions, further improving the performance. 
To obtain more fine-grained information, several works \etal~\cite{kalayeh2018human,suh2018part,gao2020pose,miao2019pose,lanmagnifiernet} introduce one extra human parsing branch to provide part matching information in the feature level. 
Some pioneering works also explore the neural architecture search to learn fine-grained visual representation~\cite{zhou2021attention,quan2019auto,ren2021comprehensive}. Besides, to address the misdetection of the input image, Zheng \etal~\cite{zheng2017pedestrian} apply the spatial transformer network~\cite{jaderberg2015spatial} to re-align feature maps. 
Different from existing works on part alignment in 2D space, the proposed method explores the 3D body structure, which is more close to the prior knowledge of human - a 3D non-rigid object. 

\subsection{Learning from Synthetic Data}
Another active research line is to leverage the synthetic human data. Although most datasets~\cite{zheng2015scalable,ristani2016performance} provide more training data in recent years, the number of images per person is still limited~\cite{zheng2017unlabeled}. Therefore, the intra-class variants of every training pedestrian are limited, which largely compromise the model learning and hurt the model scalability to the real-world scenario. To address the data limitation, one line of existing works leverages the generative adversarial network (GAN)~\cite{goodfellow2014generative} to synthesize more high-quality training images, and let the model ``see'' more appearance variants to learn the robust representation \cite{zheng2019joint,ge2018fd,eom2019learning,qian2018pose,zheng2017unlabeled,liu2018pose,zhong2018camera,zou2020joint}. Zheng \etal~\cite{zheng2017unlabeled} first propose a new label smooth regularization for outliers to leverage imperfect generated images. In a similar spirit, Huang \etal~\cite{huang2018multi} deploy the pseudo label learning to assign refined labels for synthetic data. Qian \etal~\cite{qian2018pose} modify the generation model and add pedestrian images with different poses into training set, yielding the pose-invariant features. 
Inspired by the conventional encoder-decoder manner, Ge \etal~\cite{ge2018fd} propose FD-GAN to learn one pose-invariant feature when encoding the input image. DG-Net~\cite{zheng2019joint} disentangles the pedestrian image to two embeddings, \ie, appearance code and structure code, to generate diverse and realistic synthetic images. With the high-quality synthetic data, more discriminative feature can be learned, in turn, improving re-id performance. 
Furthermore, several works ~\cite{deng2018image,zhong2018camera,zhong2020learning,wang2019learning,wei2018person} also apply GAN, \ie, CycleGAN~\cite{CycleGAN2017}, to cross-domain person re-identification by training the model with the target-style synthetic data.
In contrast, another line of works \cite{sun2019dissecting,tang2019pamtri,yao2019simulating} is close to our work, which applies the game engine to build 3D models. Sun \etal~\cite{sun2019dissecting} build a large number of 3D person models, and map models to 2D plane for generating more 2D training data. Yao \etal~\cite{yao2019simulating} and Tang \etal~\cite{tang2019pamtri} manipulate the generation setting and leverage attributes, \eg, color and pose, to enable multi-task learning on 2D synthetic data. 
Lin \etal~\cite{lin2020cross} also leverage the synthetic data to learn the common knowledge of human structure, improving the model scalability on real data. However, different from our work, the above-mentioned studies are mostly investigated in the 2D space, and neglect the 3D geometry information of human bodies. In this work, we argue that the 3D space with the geometry knowledge could help to learn a new type of feature free from several intra-class visual variants, such as viewpoints.

\subsection{Learning from Point Clouds} 
The point cloud is a flexible geometric representation of 3D data structure, which could be obtained by most 3D data acquisition devices, such as radar. The point cloud data is usually unordered, and thus the conventional convolutional neural network (CNN) could not directly work on this kind of data. One of the earliest works, \ie, PointNet \cite{qi2017pointnet}, proposes to leverage the multi-layer perceptron (MLP) networks and max-pooling layer to fuse the information from multiple points. PointNet++ \cite{qi2017pointnet++} takes one more step by introducing the sampling layer to distill salient points. To address the limitation in decoding, FoldingNet~\cite{yang2018foldingnet} adds one constant 2D plane to simulate the surface of 3D objects. However, the communication between the 3D points is still limited, and each point is treated independently most of the time. Therefore, Wang \etal~\cite{wang2019dynamic} propose to leverage Graph Neural Network (GNN) \cite{scarselli2008graph} to enable the information spread between the $k$-nearest points. Li \etal \cite{li2019deepgcns} take one more step and propose to deploy a deeper graph neural network structure, further boosting the performance.
Similarly, in this work, we regard every person as one individual graph, while every RGB pixel and the corresponding location are viewed as one node in the graph. More details are provided in Section \ref{sec:method}.   

\begin{figure*}[t]
\begin{center}
     \includegraphics[width=1\linewidth]{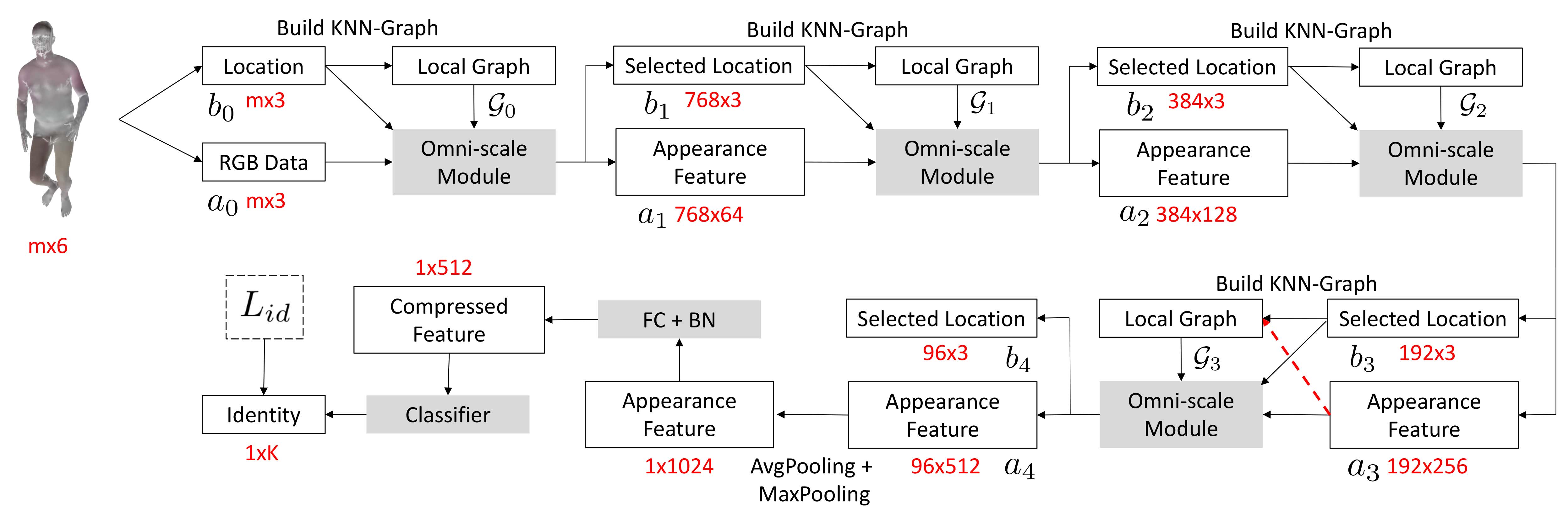}
\end{center} 
      \caption{ \textbf{OG-Net Architecture}. OG-Net is simply built via stacking Omni-scale Modules. $(m'\times c)$ denotes the feature of $m'$ points with $c$-dim attribute. Given the point cloud of $(m\times 6)$, we split the geometry location $b_0$ and the rgb color data $a_0$. The 3D location information, \ie, (x,y,z), is to build the KNN graph, while the rgb data is to extract the appearance feature as the conventional 2D CNNs. We progressively downsample the number of selected points $\{m,768,384,192,96\}$, while increasing the appearance feature length $\{3,64,128,256,512\}$. For the last KNN Graph, we concatenate the position $b_3$ and the appearance feature $a_3$ to yield a non-local attention (see the red dash arrow). Finally, we concatenate the outputs of average pooling and max pooling layer, followed by one fully connected (FC) layer and one batch normalization (BN) layer. We adopt the conventional pretext task, \ie, identity classification $
     L_{id}$, as the optimization objective to learn the pedestrian representation. When testing, we drop the last classifier and extract the compressed feature of $512$ dimensions as the pedestrian representation for matching. }
      \label{fig:Framework}
\end{figure*}

\section{Method} \label{sec:method}
We show a schematic overview of our framework in Figure~\ref{fig:Framework}. We next introduce some notations and assumptions, followed by the details of how to learn from 3D points, and how to take advantage of 2D appearance information and 3D structure in one coherent manner. 

\subsection{Preliminaries and Notations}
To conduct person re-identification in the 3D space, we first change the data structure of inputs. In particular, given one person re-id dataset, 2D images are mapped to the 3D space via the off-the-shelf 3D pose estimation \cite{kanazawaHMR18}. We apply this mapping function to every image in the dataset to obtain 3D point clouds aligned with the 2D appearance. 
We denote the generated point sets and identity labels as $S=\{s_n\}_{n=1}^N$ and $Y=\{y_n\}_{n=1}^N$, where $N$ is the number of samples in the dataset, $y_n \in [1,K]$, and $K$ is the number of the identity categories. 
We utilize the matrix format to illustrate the point cloud  $s_n \in \mathbb{R}^{m \times 6}$, where $m$ is the number of points, and 6 is the channel number. The former $3$ channels contain 3D coordinates XYZ, while the latter $3$ channels contain the corresponding RGB information. 
Given one 3D data  $s_n \in \mathbb{R}^{m \times 6}$, our work intend to learn a mapping function $F$ which projects the input $s_n$ to the identity-aware representation $f_n=F_{\Theta}(s_n)$ with learnable parameters $\Theta$. 
Unlike the conventional image format, the 3D point clouds are unordered and discrete. We can not directly apply the traditional 2D convolutional layer on $m\times6$ to capture the local information, \eg, one $3\times 3$ receptive field, since unordered neighbor points may have limited connections to the center point. To address the limitation, we follow the idea of graph neural networks~\cite{scarselli2008graph} to build the graph $\mathcal{G}$ based on the distance between points. Next we illustrate one basic component, \ie, dynamic graph convolution, to learn from the graph $\mathcal{G}$. 

\begin{figure*}[t]
  \begin{center}
    \includegraphics[width=1\linewidth]{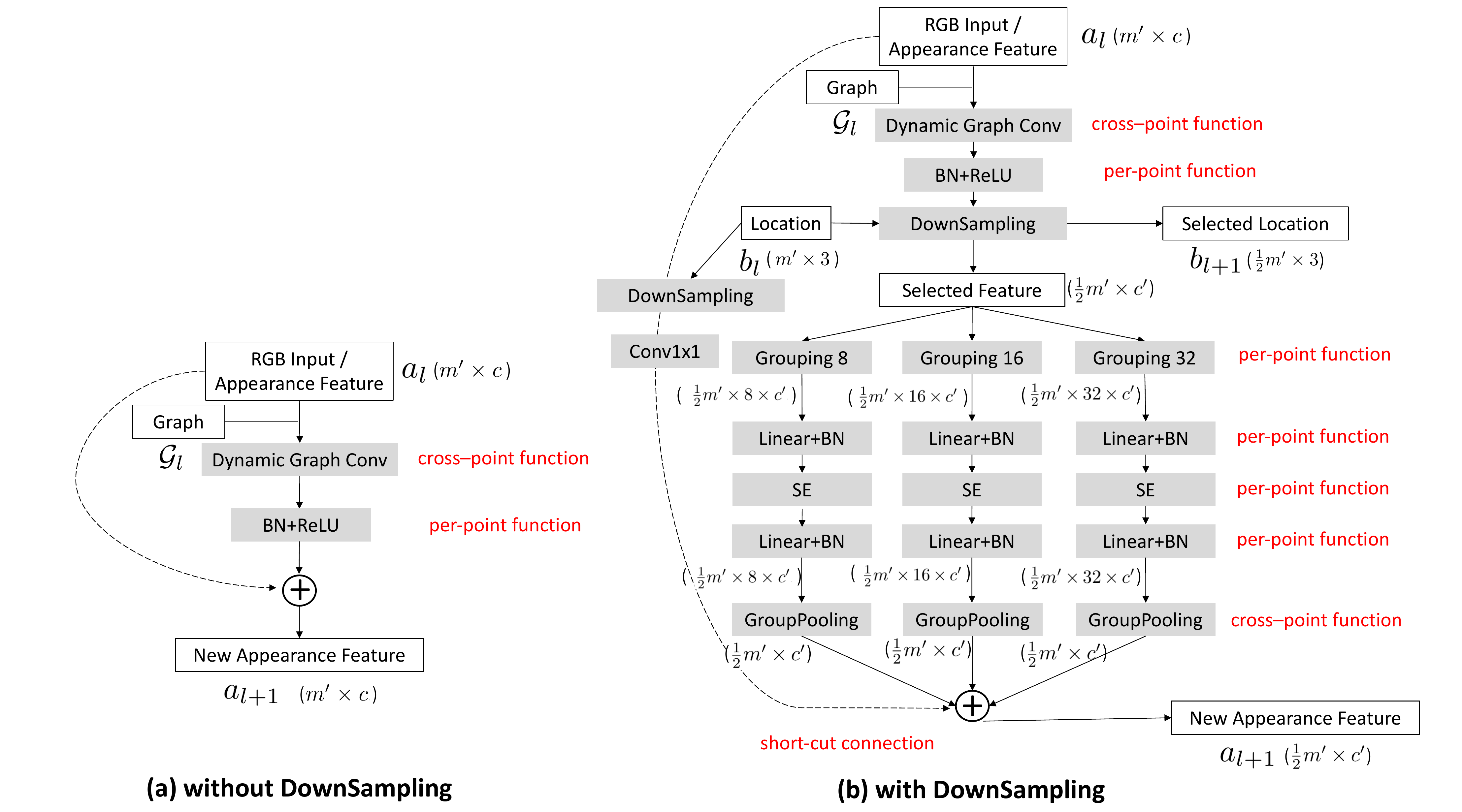}
  \end{center}  
  \caption{Visualization of Omni-scale Module. We provide the feature shape as the format of $(\cdot)$. For instance, $(m'\times c)$ denotes the feature of $m'$ points with $c$-dim attribute. (a) We show the basic Omni-scale module without downsampling. (b) We show the Omni-scale module with downsampling, which is similar to the conventional pooling layer. The module distills the number of the points and improves the training efficiency. 
  The dash line denotes the short-cut connection. 
  Besides, we highlight two function types, \ie, cross-point functions and per-point functions, in \textcolor{red}{red}. The cross-point function aggregates the feature among neighbor points, while the per-point function only considers the single-point feature. The proposed Omni-scale module consists of these two kinds of functions.}
  \label{fig:module}
\end{figure*}
 
\subsection{Dynamic Graph Convolution} 
To model the relationship between neighbor points, we adopt the $k$-nearest neighbor (KNN) graph $\mathcal{G}=(\mathcal{V},\mathcal{E})$, where $\mathcal{V}$ denotes the vertex set, and $\mathcal{E}$ denotes the edge set ($\mathcal{E}\subseteq \mathcal{V} \times \mathcal{V}$). The KNN graph is directed, and includes self-loop, meaning $(i,i)\in \mathcal{E}$. It is worth to noticing that the selection of the  $k$-nearest neighbors is based on the value of vertexes (points) rather than the initial input order, evading the problem of unordered 3D point clouds. 
Besides, recent works~\cite{wang2019dynamic,li2019deepgcns} also show the dynamic graph is superior to the fixed graph structure during training GCN, which alleviates the over-smoothing problem and enlarges the receptive field of every node. 
Following the spirit of the dynamic graph, the KNN graph used in our work is not fixed, and we re-build the graph after every down-sampling layer. The down-sampling layers are to progressively remove redundant points (vertexes), and thus the computation cost of the proposed method is much less than the conventional implementation in ~\cite{wang2019dynamic,li2019deepgcns}. 

To learn representation from the topology structure of the graph, we follow the spirit of the traditional 2D CNN and deploy one local convolutional layer based on neighbor points with connected edges. In particular, given one node feature $x_i$, the output $x'_i$ of the dynamic graph convolution could be formulated as: 
\begin{equation}
    x'_i = \sum_{j:(i,j)\in \mathcal{E},~j\neq i}(\theta_i x_i + \theta_j x_j) 
\end{equation}
where $x_j$ is the feature of neighbor points in the graph, and there is one edge from $i$ to $j$. $\theta$ is the learnable parameter in $\Theta$. The main difference with the traditional convolution is the definition of the neighbor set. \textbf{In this work, we combine two kinds of neighbor choices, \ie, position similarity and feature similarity.} If the graph $\mathcal{G}$ is based on the 3D coordinate similarity, dynamic graph convolution equals to the conventional 2D CNN to capture the local pattern based on the position. We note that this operation is translation invariant, since the global translation, such as ShiftX, ShiftY and Rotation, could not change the connected neighbors in $\mathcal{E}$.
On the other hand, if the graph $\mathcal{G}$ is built on the appearance feature, the dynamic graph convolution works as the non-local self-attention as \cite{wang2018non,zhang2018self}, which ignores the local position but pays attention to the area with similar appearance patterns. We next take advantage of the dynamic graph convolution function to build the basic module - Omni-scale module. 

\subsection{Omni-scale Module}
To leverage the rich multi-scale information as the prevailing 2D CNNs, we propose one basic Omni-scale module, which could be easily stacked to form the whole network. The module treats the 3D location and the RGB input differently (see Figure~\ref{fig:module}~(b)). 
We denote $l \in[0,L-1]$ as the layer index. The RGB input is the first appearance feature $a_0$ of $m \times 3$, while the initial 3D position is $b_0$ of $m \times 3$. Different from the conventional graph CNN, the local $k$-nearest graph $\mathcal{G}_l$ is dynamically generated according to the input location $b_l$ or the concatenation of $a_l$ and $b_l$. Given the appearance feature $a_l$ of $m' \times c $, the location $b_l$ of $m' \times 3$ and the KNN graph $\mathcal{G}_l$, the Omni-scale module outputs the appearance feature $a_{l+1}$ and the selected locations $b_{l+1}$. 
From the top to the bottom of the module, we first apply Dynamic Graph Convolution to aggregate the $k$-nearest neighbor features, which is similar to the conventional convolutional layer. Dynamic Graph Convolution does not change the number of points, and thus the shape of the output feature is $m' \times c'$. 
If down-sampling points is not applied, we will remain the channel number $c' = c$ following the conventional residual learning~\cite{he2016deep}  to obtain $a_{l+1}$ followed by one batch normalization layer and one ReLU (see Figure~\ref{fig:module}~(a)). 
If down-sampling points is applied, we generally set $c' = 2c$ to enlarge the feature channel before downsampling. Then we downsample the location according to the farthest point sampling (FPS) \cite{qi2017pointnet++}. FPS selects the most distinguish points in the 3D space. We note that only the 3D position $b_l$ is used to calculate the distance and decide the selected points when downsampling. According to the selected location, we also downsample the appearance feature, and only keep the feature of the selected location. Therefore, the shape of the selected location is $ \frac{1}{2} m' \times 3$, while the selected feature shape is $ \frac{1}{2} m' \times c'$. 
Next we deploy three branches with different grouping rates $r=\{8, 16, 32\}$, and the three branches do not share weights. In this way, we could capture the information with different receptive fields as the conventional 2D CNNs, \ie, InceptionNet~\cite{szegedy2017inception}. Each branch consists one grouping layer, two linear layers, two batch normalization (BN) layers, one squeeze-excitation (SE) block~\cite{hu2018squeeze} and one group max pooling layer to aggregate the local information. 
Specifically, grouping-$r$ layer is to sample and duplicate the $r$ nearest points for each point, followed by the linear layers, batch normalization and the SE block. We introduce SE-block~\cite{hu2018squeeze} as one adaptive gate function to re-scale the weight of each branch 
before the summarization of three branches. Group max pooling layer is to maximize the feature within each group. 
Finally, we adopt the `add' to calculate the sum of three branches rather than concatenation, so that the different scale pattern of the same part, such as cloth logos, could be accumulated. The shape of the new appearance feature $a_{l+1}$ is $ \frac{1}{2} m' \times c'$, and the shape of the corresponding 3D position $b_{l+1}$ is $ \frac{1}{2} m' \times 3$. Alternatively, we could add the short-cut connection to take advantage of the identity representation as ResNet~\cite{he2016deep}. 

To summarize, the key of Omni-scale Module is two cross-point functions. The cross-point function indicates the function considers the neighbor points, while the pre-point function only considers the feature of one point itself. One cross-point function is the dynamic graph convolution before downsampling, which could be simply formulated as $\sum{h(x_i, x_j)}$, where $h$ denotes a linear function. It mimics the conventional 2D CNN to aggregate the local patterns according to the position.
The other is the max group pooling layer in each branch, which could be simply formulated as $\max{h(x_i)}$. It maximizes neighbor features in each group as the new point feature. 
Now we have the Omni-scale module to learn from both of the appearance and the geometry structure information in a coherent manner, and next we will utilize Omni-scale modules to build the Omni-scale Graph Network (OG-Net).

\subsection{OG-Net Architecture} 
The structure of OG-Net is as shown in Figure~\ref{fig:Framework}, consisting four Omni-scale modules. We progressively decrease the number of selected points as the conventional CNN. Every time the point number decreases, the channel number of the appearance feature is doubled. After four Omni-scale modules, we could obtain $96$ points with $512$-dim appearance feature. Similar to \cite{wang2019dynamic}, we apply the max pooling as well as average pooling to aggregate the point feature, and concatenate the two outputs, yielding the $1024$-dim feature. We add one fully-connected layer and one batch normalization layer to compress the feature to $512$ dimensions as the pedestrian representation. When inference, we drop the last linear classifier for the pretext classification task, and extract the $512$-dim feature to conduct image matching. 

\noindent\textbf{Training Objective.} 
We adopt the conventional identity classification as the pretext task to learn the identity-aware feature. The vanilla cross-entropy loss could be formulated as:
 \begin{equation}
     L_{id} = \mathbb{E}[-log(p(y_n|s_n))]
 \end{equation}
where $p(y_n|s_n)$ is the predicted possibility of $s_n$ belonging to the ground-truth class $y_n$. The training objective demands that the model could discriminate different identities according to the input points. Besides, other training objectives are orthogonal to our work. 
1) In this work, we intend to show the strong potential ability of the 3D space and the proposed OG-Net. We, therefore, only deploy the basic identification loss for a fair comparison with other networks.
2) We deploy the new-released circle loss~\cite{sun2020circle} to show that our work can be fused with better loss functions for further performance boost.

\noindent\textbf{Relation to Existing Methods.} 
The main difference with existing GNN-based networks~\cite{wang2019dynamic,yang2018foldingnet} is three-fold: (1) We extract the multi-scale local information via the proposed Omni-scale Block, which can deal with the common scale variants in 3D person data; (2) We split the XYZ position information and RGB color information, and treat them differently. RGB inputs are used to extract appearance features, while the geometry position is to build the graph for local representation learning; (3) Due to a large number of points in 3D person, we progressively reduce the number of nodes in the graph, facilitating efficient training for 3D person data. 
On the other hand, compared with PointNet~\cite{qi2017pointnet} and PointNet++ \cite{qi2017pointnet++}, the proposed OG-Net contains more cross-point functions, and provides topology information, enriching the representation power of the network. The graph could be built on the two kinds of neighbor choices, \ie, position similarity or feature similarity.

\section{Experiment}
\subsection{Implementation Details}  
OG-Net is trained with a mini-batch of $36$. We deploy Adam optimizer~\cite{kingma2014adam} with amsgrad~\cite{reddi2019convergence} and the initial learning rate is set to $8e-4$. We gradually decrease the learning rate via the cosine policy~\cite{loshchilov2016sgdr}, and the model is trained for $1000$ epochs. To regularize the training, we transfer some traditional 2D data augmentation methods, such as random scale and position jittering, to the 3D space. For instance, position jittering is to add zero-mean Gaussian noise to every point. Following the setting in DGCNN~\cite{wang2019dynamic}, we set the neighbor number of KNN-graph to $k=20$. The dynamic graph convolution in OG-Net can be any of the existing graph convolution operations, such as EdgeConv~\cite{wang2019dynamic}, SAGE~\cite{hamilton2017inductive} and GAT~\cite{xu2018powerful}. In practise, we adopt EdgeConv~\cite{wang2019dynamic}. Dropout with 0.7 drop probability is used before the last linear classification layer. Since the basic OG-Net is shallow, we do not use the short-cut connection. For the person re-id task, the input image is resized to $128\times64$, and there are $8192$ points with RGB color information. After mapping to the 3D space, we uniformly sample half points to train the OG-Net, and thus the number of input $m$ in Figure~\ref{fig:Framework} equals to 4096. We note that, for other competitive 2D CNN methods, we still follow the common setting, and the 2D image input is resized to $256\times128$ ~\cite{sun2017beyond,zheng2019joint} for a fair comparison.  

\noindent\textbf{OG-Net.} The channel number of the four Omni-scale Module in OG-Net is \{64, 128, 256, 512\}. The parameter number is \textbf{$1.95 M$}, which is much less than the prevailing CNN structure ResNet-50 ($24.56 M$). 

\noindent\textbf{OG-Net-Small.} To compare with lightweight models, we also introduce OG-Net-Small with fewer channel numbers, \ie, \{48, 96, 192, 384\}. The parameter number of the model is \textbf{$1.20 M$}, which is less than both widely-adopted mobile models, \ie, ShuffleNetV2 ($1.78 M$) and MobileNetV2 ($4.16 M$). 

\noindent\textbf{OG-Net-Deep.} We build one deep OG-Net with more Omni-scale Modules. The channel numbers are \{48, 96, 96, 192, 192, 384, 384\}. The short-cut connection is enabled. Further discussion on short-cut connection is provided in Table~\ref{table:res}.  The parameter number is $2.47 M$. 

The models are trained from scratch on 3D point clouds. The whole training process costs about 2 days, with one NVIDIA 2080Ti. During testing, we extract the 512-dim feature before the classifier as the pedestrian representation. The feature is L2-normalized. Given one query image, we calculate the cosine similarity between the query feature and the candidate features of gallery images. We sort gallery images and return the ranking list according to the cosine similarity. 

\subsection{Datasets} 
We verify the effectiveness of the proposed method on four large-scale person re-id datasets, \ie, Market-1501~\cite{zheng2015scalable}, DukeMTMC-reID~\cite{ristani2016performance,zheng2017unlabeled},  MSMT-17~\cite{wei2018person}, and CUHK03-NP~\cite{li2014deepreid,zhong2017re}. 

\noindent\textbf{Market-1501}~\cite{zheng2015scalable} is collected in a university campus by $6$ cameras, containing $12,936$ training images of 751 identities, $3,368$ query images and $19,732$ gallery images of the other $750$ identities. There are no overlapping identities (classes) between the training and test set. Every identity in the training set
has 17.2 photos on average. All images are automatically detected by the DPM detector~\cite{felzenszwalb2009object}.

\noindent\textbf{DukeMTMC-reID}~\cite{ristani2016performance,zheng2017unlabeled} consists $16,522$ training images of $702$ identites, $2,228$ query images of the other $702$ identities and $17,661$ gallery images, which is mostly collected in winter by eight high-resolution cameras. It is challenging in that most pedestrians are in the similar clothes, and may be occluded by cars or trees.

\noindent\textbf{MSMT-17}~\cite{wei2018person} is one of the newly-released large-scale datasets, including $126,441$ images collected in both indoor and outdoor scenarios with $15$ cameras. It contains $32,621$ images of $1,041$ identities for training, $11,659$ query images with $82,161$ gallery images.

\noindent\textbf{CUHK03-NP}~\cite{li2014deepreid} is one of the early person re-identification datasets. We follow the new protocol in ~\cite{zhong2017re} to split $767$ identities as the training set, and the rest $700$ identities are deployed to verify the model. We utilize the pedestrian images detected by DPM~\cite{felzenszwalb2009object} for training and testing, which is more close to the real-world scenario.
 
\noindent\textbf{Evaluation Metrics.} We report Rank-1 accuracy (R@1) and mean average precision (mAP). Rank-$i$ denotes the probability of the true match in the  top-$i$ of the retrieval results, while AP denotes the
area under the Precision-Recall curve. The mean of the average precision (mAP) for all query images reflects the
precision and recall rate of the retrieval performance. Besides, we also provide the number of model parameters (\#params). 

\begin{figure}{
\caption{\textbf{(a,b)} Visualization of lossy compression in the 2D-to-3D mapping, which drops the body outliers, \eg, hair and dress. \textbf{(c)} We still introduce the 2D background to 3D space. }\label{fig:bg}}{%
  \includegraphics[width=0.9\textwidth]{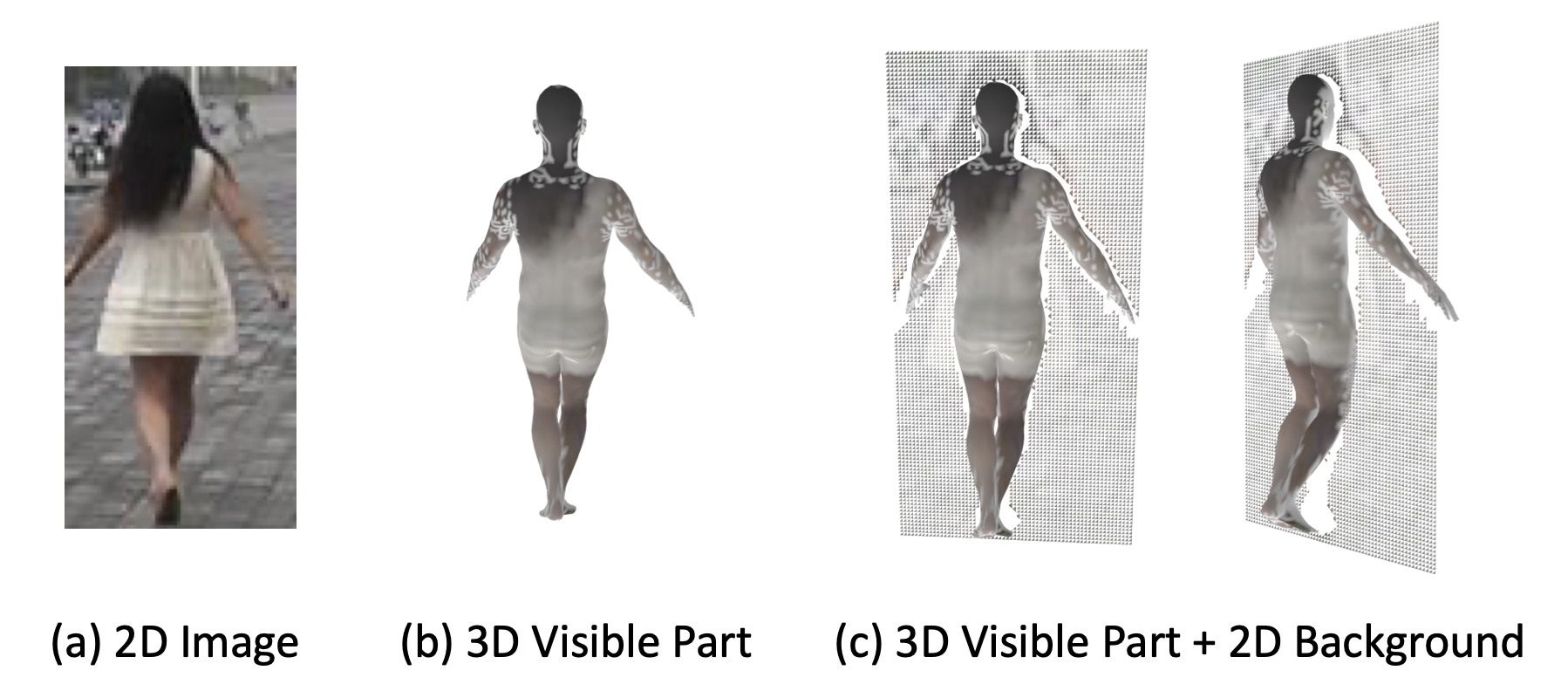}
}%
\end{figure}

\noindent\textbf{Data Limitation.} \label{sec:limitation}
Before the experimental analysis, we would like to illustrate several data limitations. It is mainly due to lossy mapping in the 2D-to-3D process.  Due to the restriction of the 3D human model, we could not build the 3D model for several body outliers, such as hair, bag, dress. However, these outliers contain discriminative identity information. For instance, as shown in Figure~\ref{fig:bg}~(a) and (b), the 3D model based on the visible part drops some part of hair and dress of the girl, which is not ideal for representation learning. We think it could be solved via the depth estimation devices, such as Kinect~\cite{han2013enhanced}, or more sophisticated human models in the future. In this paper, we do not solve the 3D human reconstruction problem, but focus on the person re-identification task. Therefore, as a trade-off, we still introduce the 2D background, and project the corresponding pixel to the XY plane (see Figure~\ref{fig:bg}~(c)).

\begin{table}[tbp]
\centering
{
\caption{Ablation study of different inputs on Market-1501. $\dagger$: For a fair comparison, the model is trained on the traditional 2D image inputs with extra 3D coordinates $(\mathbf{x},\mathbf{y},\mathbf{0})$. 
}\label{table:inputs} }{
\scalebox{0.8}{
\begin{tabular}{l|cc|cc}
\shline
\multirow{2}*{Inputs} & \multicolumn{2}{c|}{Market-1501} & \multicolumn{2}{c}{DukeMTMC-reID} \\
 & R@1 & mAP & R@1 & mAP\\
\shline
\multirow{2}*{3D Visible Part} & \multirow{2}*{77.64} & \multirow{2}*{54.52} & \multirow{2}*{59.52} & \multirow{2}*{37.25} \\
& & & &  \\
\multirow{2}*{2D Image$^\dagger$}& \multirow{2}*{85.72} & \multirow{2}*{67.28} & \multirow{2}*{75.49} & \multirow{2}*{55.98} \\
& & & & \\
\hline
\hline
\multirow{2}*{3D Visible Part + 2D Background} & \multirow{2}*{\textbf{86.79}} & \multirow{2}*{\textbf{67.92}} 
& \multirow{2}*{\textbf{77.33}} & \multirow{2}*{\textbf{57.74}} \\
& & & &  \\
\shline
\end{tabular}}
}
\end{table}

\setlength{\tabcolsep}{7pt}
\begin{table*}
\scriptsize
\caption{We mainly compare three groups of models trained from scratch on four large-scale person re-id datasets, \ie, Market-1501~\cite{zheng2015scalable}, DukeMTMC-reID~\cite{ristani2016performance,zheng2017unlabeled}, MSMT-17~\cite{wei2018person} and CUHK03-NP~\cite{li2014deepreid,zhong2017re}. We report Rank1($\%$), mAP($\%$) and the number of model paramters (M). 
The first group contains  the point-based methods that we re-implemented. The second group contains the lightweight CNN models. The third group contains prevailing 2D CNN models with more parameters. }
\begin{center}
\label{table:supervised}
\begin{tabular}{l|c|c|c|cc|cc|cc|cc}
\shline
\multirow{2}{*}{Method} & Input  & Loss &\multirow{2}{*}{\#params(M)} 
& \multicolumn{2}{c|}{Market-1501} & \multicolumn{2}{c|}{DukeMTMC-reID} & \multicolumn{2}{c|}{MSMT-17} & \multicolumn{2}{c}{CUHK03-NP}\\
& Type & Function & & R@1 & mAP & R@1 & mAP & R@1 & mAP & R@1 & mAP \\
\shline

DGCNN ~\cite{wang2019dynamic} & point clouds  & CE & 1.37 
& 28.89 & 13.33 & 29.17 & 15.16 & 2.84 & 1.19 & 3.4 & 3.6 \\
PointNet++ (SSG)~\cite{qi2017pointnet++} & point clouds & CE & 1.59 
& 61.79 & 37.89 & 55.70 & 35.16 & 22.94 & 9.61 & 14.57 & 13.97 \\
PointNet++ (MSG)~\cite{qi2017pointnet++} & point clouds & CE & 1.87 
& 72.51 & 47.21 & 60.23 & 39.36 & 28.99 & 12.52 & 21.14 & 19.79\\
PointNet++ (MSG)~\cite{qi2017pointnet++} & point clouds & CE + Circle & 1.87 
& 76.04 & 52.44 & 64.23 & 44.19 & 22.57 & 9.55 & 21.36 & 19.86 \\
\hline
ShuffleNetV2~\cite{zhang2018shufflenet} & images & CE & 1.78 
& 79.75 & 56.80 & 68.81 & 48.09 & 36.80 & 15.70 & 25.29 & 22.90 \\
ShuffleNetV2~\cite{zhang2018shufflenet} & images & CE + Circle & 1.78 & 79.78 & 58.50 & 69.34 & 49.04 & 33.16 & 14.00 & 25.43& 23.56\\
MobileNetV2~\cite{sandler2018mobilenetv2} & images & CE & 4.16 
& 81.95 & 59.28 & 71.05 & 50.45 & 42.53 & 18.62 & 29.57 & 26.45 \\
MobileNetV2~\cite{sandler2018mobilenetv2} & images & CE + Circle & 4.16 & 79.39 & 57.40 & 69.70 & 49.75 & 29.19 & 11.79 & 29.14 & 25.46 \\ 
\hline
OG-Net-Small & point clouds & CE & 1.20 
& 86.79 & 67.92 & 77.33 & 57.74 & 42.44 & 20.31 & 43.07 & 38.06 \\ 
OG-Net-Small & point clouds & CE + Circle & 1.20
& 87.38 & 70.48 & 77.15 & 58.51 & 43.84 & 21.79 & 46.43 & 41.79\\
OG-Net & point clouds  & CE & 1.95 
& 86.82 & 69.02 & 76.53 & 57.92 & 44.27 & 21.57 & 44.00 & 39.28 
 \\
OG-Net & point clouds & CE + Circle & 1.95 & 
87.80 & 70.56 & 78.37 & 60.07 & 45.28 & 22.81 & 48.29 & 43.73 \\
\shline
DenseNet-121~\cite{Huang2017Densely} & images & CE & 8.50 
& 83.14 & 63.36 & 73.16 & 55.08 & 46.32 & 21.50 & 33.64 & 29.45\\
DenseNet-121~\cite{Huang2017Densely} & images & CE + Circle & 8.50 & 84.26 & 65.79 & 74.28 & 55.75 & 41.06 & 18.46 & 36.21 & 33.52 \\
ResNet-50~\cite{he2016deep} & images & CE & 24.56 
& 84.59 & 65.31 & 73.20 & 55.96 & 46.88 & 22.25 & 35.43 & 32.09\\
ResNet-50~\cite{he2016deep} & images & CE + Circle & 24.56 & 85.27 & 67.55 & 74.15 & 56.83 & 37.35 & 16.98 & 37.29 & 34.12 \\
\hline
OG-Net-Deep & point clouds  & CE & 2.47 
& 88.36 & 71.27 &  76.97 & 59.23 & 44.56 & 21.41 & 45.71 & 41.15\\
OG-Net-Deep & point clouds  & CE+Circle & 2.47 & \textbf{88.81} & \textbf{72.91} & \textbf{78.50} & \textbf{60.70} & \textbf{47.32} & \textbf{24.07} & \textbf{49.43} & \textbf{45.71} \\
\shline
\end{tabular}
\end{center}
\end{table*}

\subsection{Quantitative Results}

\noindent\textbf{Comparisons to the 2D Space.} We compare the results on three kinds of inputs, \ie, 2D input, 3D Visible Part and 3D Visible Part with 2D Background. For a fair comparison, the grid of the 2D input is also transformed to the point cloud format as $(\mathbf{x}, \mathbf{y}, \mathbf{0})$, while $\mathbf{z}$ is set to $0$. We train OG-Net on three kinds of input data with the same hyper-parameters. As shown in Table \ref{table:inputs}, we observe that the retrieval result of the pure 3D Visible Part input is inferior to that of 2D Image. As discussed in Section \ref{sec:limitation}, we speculate that it is  due to the lossy 2D-to-3D mapping, which drops several discriminative parts, such as hair, dress, and carrying. 
In contrast, the 3D Visible Part + 2D Background has achieved superior performance $86.79\%$ Rank@1 and $67.92\%$ mAP to the result of 2D Image ($85.72\%$ Rank@1 and $67.28\%$ mAP), which shows that the 3D position information is complementary to 2D color information.  Similar results also can be observed on the DukeMTMC-reID dataset. The 3D information yields $+1.84\%$ Rank-1 and $+1.76\%$ mAP accuracy improvement. The 3D space could ease the matching difficulty and highlight the geometry structure. 

\noindent\textbf{Person Re-id Performance.}
We compare the proposed method with three groups of competitive methods, \ie, prevailing 2D CNN models, light-weight CNN models, and  popular point classification models. 
We note that the model pre-trained on the large-scale datasets, \eg, ImageNet~\cite{deng2009imagenet}, could yield the performance boost. For a fair comparison, models are trained from scratch with the same optimization objective, \ie, the cross-entropy loss. Since the proposed method is orthogonal to different metric learning losses, we also run experiments with the prevailing circle loss~\cite{sun2020circle}.  As shown in Table \ref{table:supervised}, we can make the following observations: 

(1) OG-Net has achieved competitive results of $69.02\%$ mAP, $57.92\%$ mAP, $21.57\%$ mAP, and $39.28\%$ mAP on four large-scale person re-id benchmarks with limited training parameters $1.95M$. The mobile OG-Net-Small of less channel width also achieves a close result only with the cross-entropy loss. 

(2) Comparing with the point-based methods, such as PointNet++~\cite{qi2017pointnet++} and DGCNN~\cite{wang2019dynamic}, both OG-Net and OG-Net-Small have surpassed this line of works by a clear margin, which validates the effectiveness of the proposed Omni-scale module in capturing multi-scale neighbor information on point clouds. 

(3) Comparing with light-weight 2D CNN models, \ie, ShuffleNetV2~\cite{zhang2018shufflenet} and MobileNetV2~\cite{sandler2018mobilenetv2}, OG-Net-Small has achieved competitive performance with fewer parameters ($1.20 M$). 

(4) We apply the same setting to train the model with Circle loss. The strong supervision mechanism of Circle loss sometime compromises the training process. The training process is quite challenging, especially when the class number largely increases in the MSMT-17 dataset. We observe that the proposed model is shallow and relatively easy to converge, so Circle loss generally works well with the proposed structure, and yields performance boost.  

(5) Comparing with prevailing 2D CNN models, \ie, ResNet-50~\cite{he2016deep} and DenseNet-121~\cite{Huang2017Densely}, the proposed OG-Net surpasses these models. Furthermore, OG-Net-Deep with deeper structure has achieved better Rank@1 and mAP accuracy. Besides, we also observe that OG-Net is more robust than 2D CNNs, when facing the unseen data. We will discuss this aspect in the following section.  


\noindent\textbf{Transferring to Unseen Datasets.} To verify the scalability of OG-Net, we train the model on dataset $A$ and directly test the model on dataset $B$ (with no adaptation), which is close to the real-world deployment. We denote the direct transfer learning protocol as $A \rightarrow B$. Three groups of related works are considered. We observe that the modern CNN models are typically over-parameterized, which is prone to over-fit the training dataset. As shown in Table~\ref{table:unseen}, both ResNet-50 and DenseNet-121 do not perform well given more parameters. The 3D point cloud-based methods are competitive to the conventional 2D methods. 
It is worth noting that the proposed OG-Net has outperformed the point-based methods as well as prevailing 2D networks. The results suggest that the proposed method has the potential to adapt one new re-id dataset of unseen environments. 

\setlength{\tabcolsep}{4pt}
\begin{table*}{
\caption{Transferring to unseen datasets. Here we directly deploy the model trained on the dataset $A$ to the unseen dataset $B$. We denote this setting as $A \rightarrow B$, which could reflect the scalability of the model in different scenarios. We observe that OGNet is generally superior to the ResNet-50 and DenseNet-121 as well as lightweight models, such as ShuffleNetV2 and MobileNetV2. 
}\label{table:unseen}
}{
\scalebox{0.75}{
\begin{tabular}{l|c|c|cc|cc|cc|cc|cc|cc}
\shline
\multirow{2}{*}{Method} & Input & Loss &  \multicolumn{2}{c|}{Market$\rightarrow$Duke} &  \multicolumn{2}{c|}{Duke$\rightarrow$Market} & \multicolumn{2}{c|}{Market$\rightarrow$MSMT} &  \multicolumn{2}{c|}{MSMT$\rightarrow$Market} & \multicolumn{2}{c|}{Duke$\rightarrow$MSMT} &  \multicolumn{2}{c}{MSMT$\rightarrow$Duke}  \\
& Type & Function & R@1 & mAP & R@1 & mAP  & R@1 & mAP  & R@1 & mAP  & R@1 & mAP  & R@1 & mAP \\
\shline
DGCNN ~\cite{wang2019dynamic} & point clouds & CE & 7.4 & 2.9 & 13.4 & 4.4 & 1.4 & 0.4 & 10.0 & 3.7 & 1.8 & 0.5 & 7.3 & 2.7 \\
PointNet++ (SSG)~\cite{qi2017pointnet++} & point clouds & CE & 18.6 & 8.4 & 28.8 & 11.3 & 3.9 & 1.2 & 32.4 & 13.3 & 5.5 & 1.7 & 29.0 & 15.4 \\
PointNet++ (MSG)~\cite{qi2017pointnet++} & point clouds & CE & 23.2 & 11.0& 32.8 & 12.6 & 5.0 & 1.5 & 30.6 & 12.9 & 6.5 & 1.9 & 24.3 & 12.4 \\
PointNet++ (MSG)~\cite{qi2017pointnet++} & point clouds & CE + Circle & 25.4 & 12.2 & 35.1 & 14.5 & 5.4 & 1.7 & 35.6 & 15.1 & 6.4 & 1.9 & 31.4 & 17.3 \\
\hline
ShuffleNetV2~\cite{zhang2018shufflenet} & images & CE & 17.2 & 7.2 & 36.4 & 13.9 & 2.8 & 0.8 & 36.5 & 14.1 & 5.8 & 1.5 & 29.3 & 15.3  \\
ShuffleNetV2~\cite{zhang2018shufflenet} & images & CE + Circle & 18.7 & 8.5 & 36.2 & 13.7 & 3.4 & 1.0 & 36.4 & 14.4 & 6.0 & 1.6 & 29.4 & 15.1 \\
MobileNetV2~\cite{sandler2018mobilenetv2} & images & CE & 16.7 & 7.1 & 34.3 & 12.4 & 3.2 & 0.9 & 35.9 & 14.2 & 5.5 & 1.4 & 30.6 & 15.4 \\
MobileNetV2~\cite{sandler2018mobilenetv2} & images & CE +Circle & 18.5 & 8.0 & 34.1 & 13.3 & 3.5 & 0.9 & 32.1 & 13.3 & 5.3 & 1.4 & 30.3 & 15.5 \\
\hline
DenseNet-121~\cite{Huang2017Densely} & images & CE & 11.7 & 5.0 & 32.7 & 11.6 & 2.9 & 0.8 & 34.2 & 13.0 & 5.3 & 1.5 & 27.8 & 13.6 \\
DenseNet-121~\cite{Huang2017Densely} & images & CE + Circle & 12.3 & 5.3 & 32.6 & 11.9 & 2.6 & 0.8 & 31.9 & 12.0 & 5.3 & 1.4 & 25.2 & 12.8 \\
ResNet-50~\cite{he2016deep} & images & CE & 12.1 & 5.2 & 34.3 & 13.5 & 2.7 & 0.7 & 34.7 & 13.5 & 5.4 & 1.5 & 28.1 & 14.4 \\
ResNet-50~\cite{he2016deep} & images & CE + Circle & 15.5 & 6.9 & 35.7 & 13.9 & 2.9 & 0.8 & 32.4 & 12.2 & 6.1 & 1.6 & 24.3 & 12.0 \\
\hline
OG-Net & point clouds & CE & \textbf{26.5} & 13.1 & 35.9 & 14.5 & \textbf{5.9} & \textbf{1.7} & \textbf{40.1} & \textbf{17.6} & \textbf{6.8} & \textbf{1.9} & 35.2 & \textbf{19.3} \\
OG-Net & point clouds & CE + Circle & 26.4 & \textbf{13.7} & \textbf{36.4} & \textbf{14.7} & 5.3 & 1.6 & 38.8 & 16.9 & 6.3 & \textbf{1.9}  & \textbf{35.3} &  \textbf{19.3} \\
\shline
\end{tabular}}
}
\end{table*}

\setlength{\tabcolsep}{8pt}
\begin{table}{
\caption{Effectiveness of different components. We compare the network variants, including squeeze-excitation (SE), the usage of KNN Graph and the last non-local attention in the model. 
}\label{table:components}}
{
\scalebox{0.9}{
\begin{tabular}{l|cccc}
\shline
Method & \multicolumn{4}{c}{Performance} \\
\shline
with Squeeze-excitation? &  & $\checkmark$ & $\checkmark$ & $\checkmark$  \\
with KNN Graph? &  & & $\checkmark$ & $\checkmark$ \\
with Last Non-local? & & & & $\checkmark$ \\
\hline
Rank@1 & 83.35 & 84.38 & 86.43 & \textbf{87.38}\\
mAP($\%$) & 64.65 & 65.50 & 69.33 & \textbf{70.48} \\
\shline
\end{tabular}}
}
\end{table}

\setlength{\tabcolsep}{13pt}
\begin{table}{
\caption{Effectiveness of the short-cut connection. We observe a similar result with~\cite{he2016deep} that the improvement from the short-cut connection is not significant on the ``shallow'' network, while it works well on the relatively deep network structure.
}\label{table:res}}
{
\scalebox{0.9}{
\begin{tabular}{l|c|cc}
\shline
 Method & Short-cut & R@1 & mAP \\
 \shline
 OG-Net & $\times$ & 86.82 & 69.02 \\
 OG-Net & $\checkmark$ & 84.00 & 65.04 \\
 \hline
 OG-Net-Deep & $\times$ & 86.28 & 68.49 \\
 OG-Net-Deep & $\checkmark$ & 88.81 & 72.91 \\
\shline
\end{tabular}}
}
\end{table}

\begin{figure*}[t]
\begin{center}
     \includegraphics[width=1\linewidth]{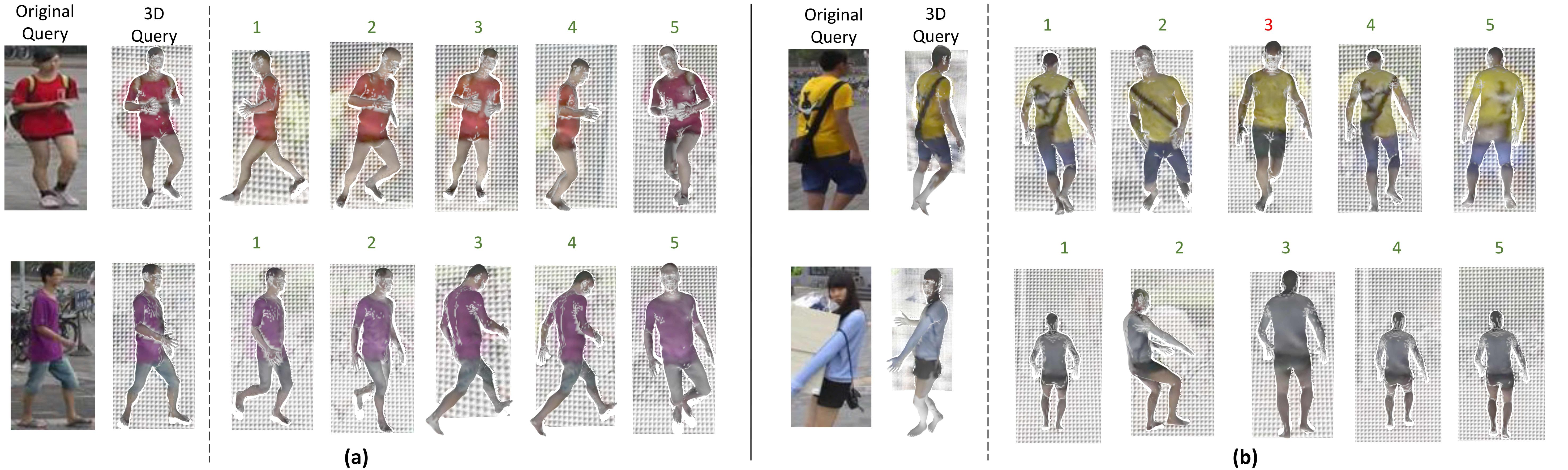}
\end{center} 
      \caption{ Visualization of Retrieval Results. \textbf{(a)} Given one 3D query, we show the original 2D images and the top-5 retrieval results. \textbf{(b)} We also show the challenging case, such as occlusion and the partially detected query. The \textcolor{OliveGreen}{green} index indicates the true-matches, while the \textcolor{red}{red} index denotes the false-matches.  }
      \label{fig:visual}
\end{figure*}

\subsection{Qualitative Results}
\noindent\textbf{Visualization of Retrieval Results.}
As shown in Figure~\ref{fig:visual}, we provide the original query, the corresponding 3D query and the top-5 retrieved candidates. Two different cases are studied. One is the typical case that the 3D human reconstruction is relatively good. OG-Net can successfully retrieve the true-matches of different viewpoints (see Figure~\ref{fig:visual} (a)). On the other hand, we also show the challenging case, including the partially detected query and occlusion. Thanks to the prior knowledge of the human geometry structure, OG-Net can still provide reasonable retrieval results with large scale variants (see Figure~\ref{fig:visual} (b)). It also verifies the robustness of the proposed approach.

\begin{figure}[t]{
\caption{Sensitivity analysis on the different number of neighbors $k$. We provide the corresponding re-id performance on Market-1501 in terms of Rank@1(\%) and mAP(\%).  }\label{fig:ablation_k}
{%
  \includegraphics[width=1\textwidth]{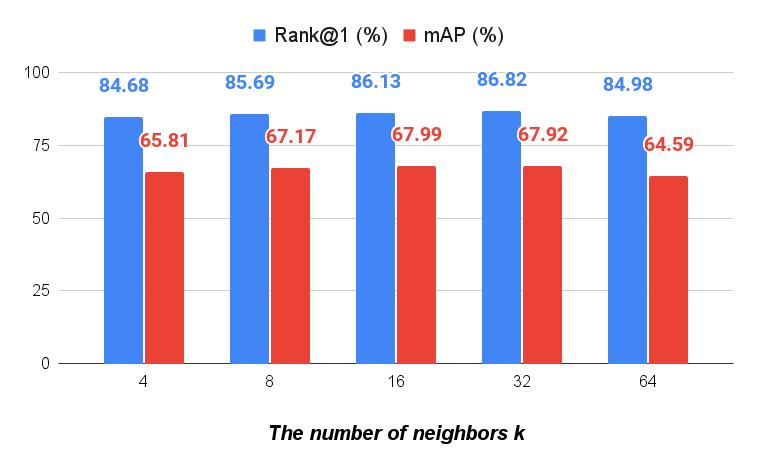}}%
}\end{figure}

\begin{figure}[t]{
\caption{Sensitivity analysis on the point density. \textbf{(left)} We visualize point clouds with different proportion of the point number. \textbf{(right)} We provide the corresponding re-id performance in terms of Rank@1(\%) and mAP(\%) against the point number variants.  }\label{fig:sample_num}
{%
  \includegraphics[width=1\textwidth]{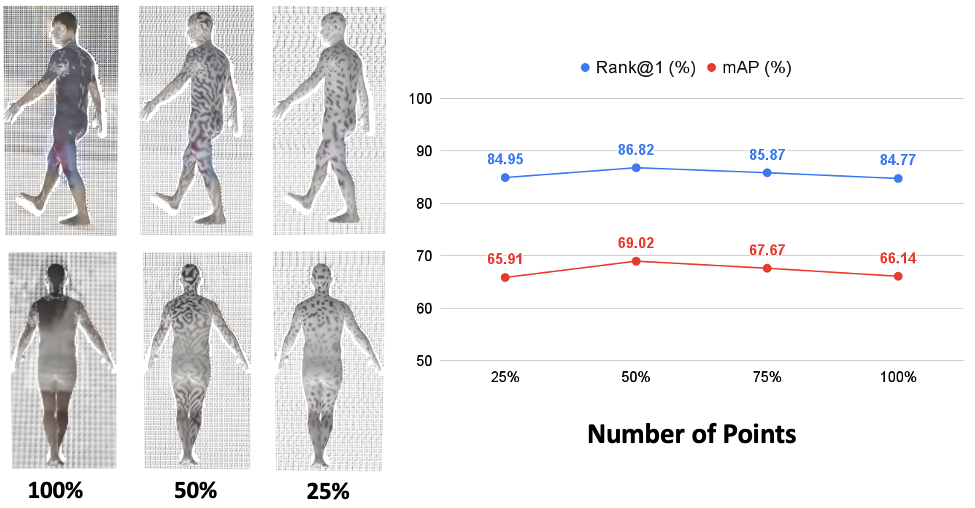}}%
}\end{figure}

\setlength{\tabcolsep}{5pt}
\begin{table}[t]{
\caption{Classification results on ModelNet~\cite{wu20153d}. We do not focus on the point cloud classification problem, but show the feasibility of the proposed OG-Net. $^\dagger$: We provide results based on our re-implementation, which is slightly higher than the reported result in \cite{qi2017pointnet++}. } \label{table:model}
{
\scalebox{0.75}{
\begin{tabular}{l|c|cc}
\shline
\multirow{2}{*}{Method} & \multirow{2}{*}{\#params(M)} & Mean-class & Overall \\
& &  Accuracy & Accuracy \\
\shline
3DShapeNets~\cite{wu20153d} & - & 77.3  & 84.7\\
VoxNet~\cite{maturana2015voxnet} & - & 83.0 & 85.9 \\
PointNet~\cite{qi2017pointnet} & 3.50 & 86.0 & 89.2 \\
SpecGCN~\cite{wang2018local} & 2.05 & - & 91.5\\
PointNet++(SSG)$^\dagger$~\cite{qi2017pointnet++} & 1.62 & 89.5 & 92.0\\
PCNN by Ext~\cite{atzmon2018point} & 1.40 & - & 92.2 \\
PointNet++(MSG)$^\dagger$~\cite{qi2017pointnet++} & 1.89 & 90.1 & 92.7 \\
DGCNN~\cite{wang2019dynamic} & 1.81 & 90.2 & 92.9 \\
Point Cloud Transformer~\cite{guo2020pct} & 2.88 & - & 93.2 \\
\hline
OG-Net-Small & \textbf{1.22} & \textbf{90.5} &  \textbf{93.3} \\
\shline
\end{tabular}}
}}
\end{table}

\section{Further Analysis and Discussions}

\noindent\textbf{Effect of Different Components.} In this section, we intend to study the mechanism of the Omni-scale Module. First, we compare the OG-Net without KNN Graph, \ie, $k=1$. For a fair comparison, we apply one linear layer to replace the dynamic graph convolution. 
As shown in the second and the third column of Table~\ref{table:components}, the performance of OG-Net without leveraging the KNN neighbor information drops from $69.33\%$ mAP to $65.50\%$ mAP. The result suggests that the dynamic graph captures effective local information, which could not be replaced by pre-point function, \eg, linear layer. On the other hand, if we include too many neighbors, \eg, $k=64$, the model loses the discriminative feature of local patterns, thus compromising the retrieval performance as well. To validate this points, we evaluate the sensitivity analysis on $k=\{4, 8, 16, 32, 64\}$ (see Figure~\ref{fig:ablation_k}). The observation is consistent with the conventional $k$ nearest neighbor algorithms~\cite{peterson2009k} on the neighbor number.

Next, we intend to verify the effectiveness of the last non-local graph. The last graph is built on the $k$-nearest neighbor of the appearance feature. (In practice, we append the 3-channel position to the appearance feature for building the graph, which prevents duplicate nodes with the same node attribute in the graph.) For a fair comparison, we replace the last non-local graph with the graph based on 3D position only. As shown in the third and the fourth column of Table~\ref{table:components}, OG-Net with the last non-local block has surpassed the model with position graph $+1.15\%$ mAP, indicating that the last non-local graph provides effective long-distance attention.  

Finally, we study two alternative components, \ie, SE block and short-cut connection. By default, Omni-scale Module deploys SE block but does not add the short-cut connection. As shown in the first and second column in Table~\ref{table:components}, we can observe that SE Block improves about $+0.85\%$ mAP from $64.65\%$ to $65.50\%$.
On the other hand, the short-cut connections do not provide significant improvement or performance drop on OG-Net, since OG-Net is relatively shallow with four Omni-scale blocks. As shown in Table~\ref{table:res}, we deploy the OG-Net-Deep to further validate this point. The observation is consistent with ResNet~\cite{he2016deep}. The short-cut connection works well on the relatively deep network structure. The performance is improved from $68.49\%$ mAP to $72.91\%$ mAP, and the short-cut connections help the model optimization. 

\noindent\textbf{Sensitivity Analysis on the Point Density.} Our model is trained with $50\%$ points, \ie, 4096, and thus the best performance is achieved with $50\%$ points remaining. In practice, different depth estimation devices may provide different scan point density. To verify the robustness of the proposed OG-Net on point density, we synthesize the data similar to that in Figure~\ref{fig:sample_num} (left) and conduct the inference. When $25\%$ points remain, OG-Net still could arrive at $84.95\%$ Rank@1 and $65.91\%$ mAP. When $100\%$ points are used, OG-Net arrives at $84.77\%$ Rank@1 and $66.14\%$ mAP. It is because too low/high density impacts the distribution of the $k$-nearest neighbors, compromising the retrieval performance. Despite the density changes, the relative  performance drop is small. The result verifies OG-Net is robust to different point density (see Figure~\ref{fig:sample_num} (right)). 

\noindent\textbf{Evaluation of Point Cloud Classification Task.} We also evaluate the proposed OG-Net on the traditional point cloud classification benchmark, \ie, ModelNet~\cite{wu20153d}. The ModelNet dataset contains 12,311 meshed CAD models of 40 categories. Following the train-test split in \cite{wang2019dynamic}, 9,843 models are used for training, while the rest 2,468 models are for evaluation. \textbf{Note that the ModelNet dataset does not provide RGB information.} To verify the effectiveness of OG-Net, we duplicate the xyz input as the appearance input to train OG-Net. Following other competitive approaches~\cite{wang2019dynamic,qi2017pointnet,qi2017pointnet++}, the number of input points is fixed as 1024. As shown in Table \ref{table:model}, we compare with prevailing models in terms of mean-class accuracy and overall accuracy. Although the proposed method is not designed for cloud point classification task, OG-Net-Small has achieved a competitive result of $90.5\%$ mean-class accuracy and $93.3\%$ overall accuracy with $1.22M$ parameters. 

\section{Conclusion}
In this work, we provide an early attempt to learn the pedestrian representation in the 3D space, easing the part matching on 2D images. The 3D assumption is aligned with the human visual system of associating the 2D appearance with the 3D geometry structure. 
Different from existing CNN-based approaches, the proposed Omni-scale Graph Network (OG-Net) takes the advantage of  3D prior knowledge and 2D appearance information in an end-to-end manner, starting from 3D human point clouds. Given 3D points and the nearest neighbour graph, the basic Omni-scale module can aggregate different-scale neighbor information in the topology, enriching the representation ability. This allows the proposed OG-Net efficiently learns discriminative feature via limited network parameters. Extensive experiments suggest that OG-Net exploits the complementary information of 3D geometry information and the 2D appearance, yielding the competitive performance on four person re-id benchmarks. The 3D prior knowledge also benefits the model generalizability on the unseen pedestrian data, which is close to the application in real-world scenarios. 

The proposed OG-Net still have room for futher improvements. 
In experiment, the proposed method learns the representation from the generated 3D point clouds mapping from 2D images. Although it works, the original 2D images are usually resized and compressed in most person re-id datasets, compromising the body shape, \eg, height. 
We may consider collecting a new 3D dataset in the future. Furthermore, the proposed method has the potential to many related fields. Similar graph-based models can be employed to other potential fields, \eg, objects with a rigid structure like vehicles~\cite{tang2019cityflow,zhang2019part} and products~\cite{wei2019rpc,liu2012hi}.  


{\footnotesize
\bibliographystyle{IEEEtran}
\bibliography{egbib}
}

\begin{IEEEbiography}[{\includegraphics[width=1in,height=1.25in,clip,keepaspectratio]{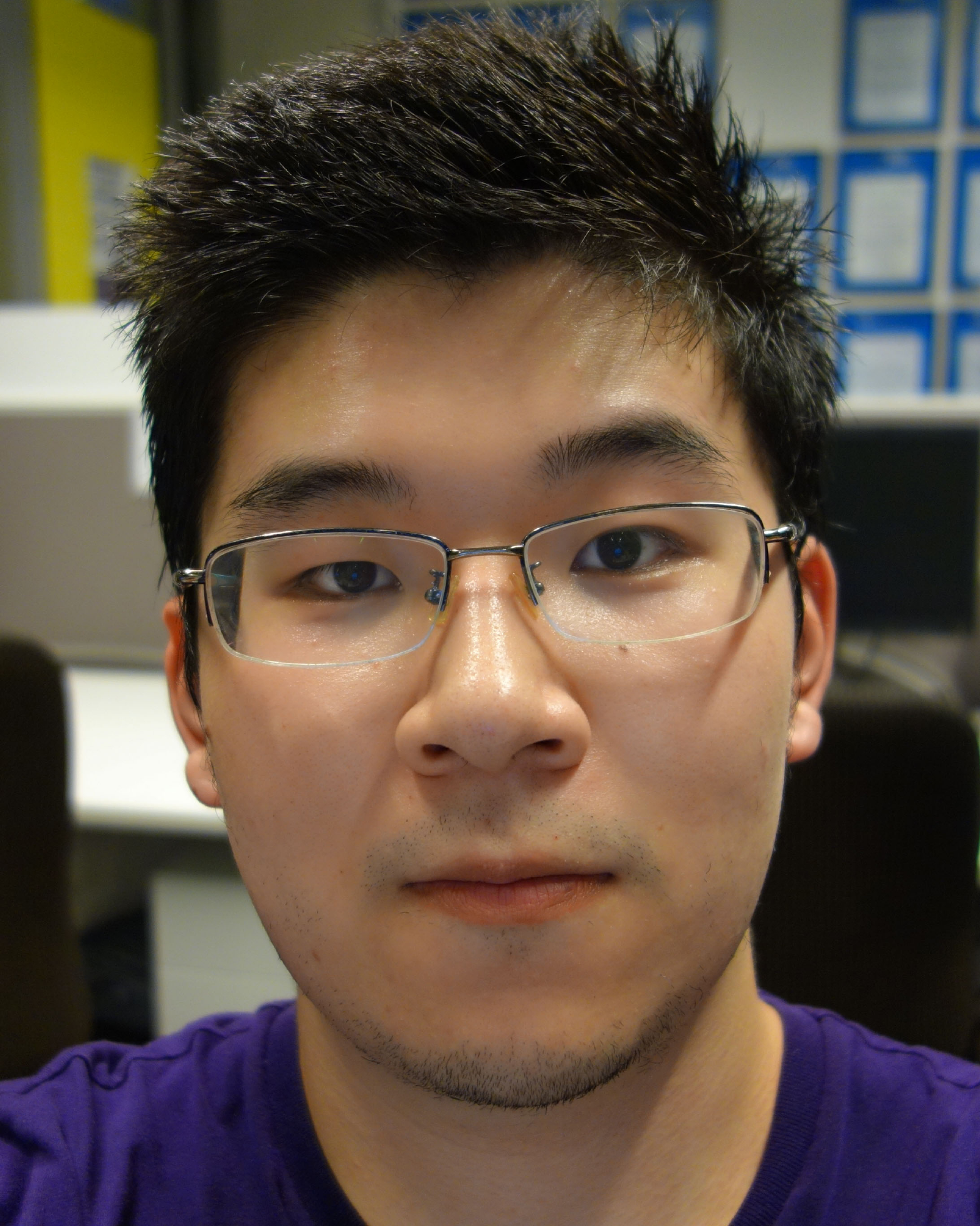}}]{Zhedong Zheng}
received the B.S. degree in computer science from Fudan University, China, in 2016. He is currently a Ph.D. student with the School of Computer Science at University of Technology Sydney, Australia. His research interests include robust learning for image retrieval, generative learning for data augmentation, and unsupervised domain adaptation.
\end{IEEEbiography}
\vfill
\begin{IEEEbiography}[{\includegraphics[width=1in,height=1.25in,clip,keepaspectratio]{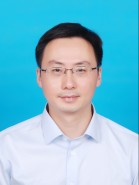}}]{Nenggan Zheng}
received the B. E. degree in Biomedical Engineering and the Ph. D. degree in Computer Science, both from Zhejiang University, in 2002 and 2009, respectively. He is currently a professor in computer science with the Academy for Advanced Studies, Zhejiang University. His current research interests include artificial intelligence, embedded systems, and brain-computer interface.
\end{IEEEbiography}
\vfill
\begin{IEEEbiography}[{\includegraphics[width=1in,height=1.25in,clip,keepaspectratio]{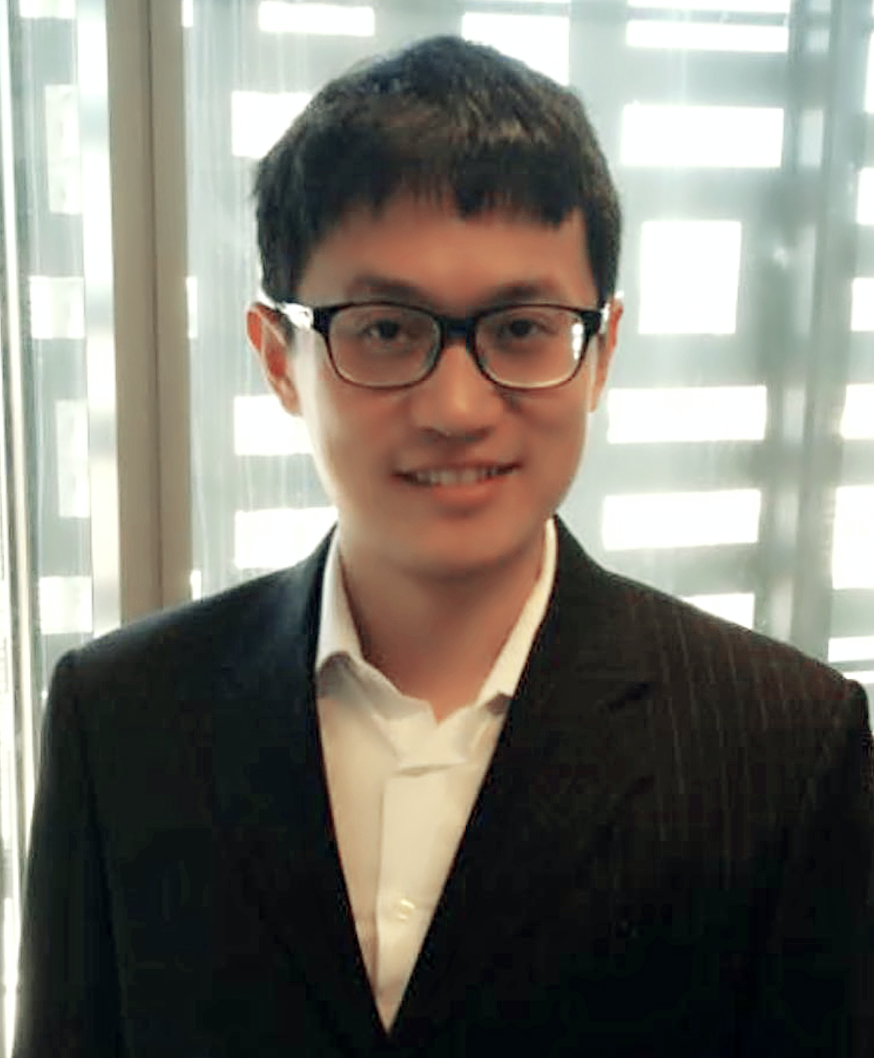}}]{Yi Yang} received the Ph.D. degree in computer
science from Zhejiang University, Hangzhou, China, in 2010. He is currently a professor with University of Technology Sydney, Australia.
He was a Post-Doctoral Research with the School of Computer Science, Carnegie Mellon University, Pittsburgh, PA, USA. His current research interest includes machine learning and its applications to multimedia content analysis and computer vision, such as multimedia indexing and retrieval, video analysis and video semantics understanding.
\end{IEEEbiography}

\end{document}